\documentclass[conference]{IEEEtran}
\usepackage{cite}
\usepackage{amsmath,amssymb,amsfonts}
\usepackage{algorithmic}
\usepackage{graphicx}
\usepackage{textcomp}
\usepackage[utf8]{inputenc}
\usepackage{rotating}
\usepackage{verbatim}
\usepackage{array}
\usepackage{hyperref}
\usepackage{ulem}
\usepackage[utf8]{inputenc}
\usepackage{multirow}
\usepackage{placeins}
\usepackage{xcolor}
\usepackage{pbalance}
\hypersetup{hidelinks}

\def\BibTeX{{\rm B\kern-.05em{\sc i\kern-.025em b}\kern-.08em
    T\kern-.1667em\lower.7ex\hbox{E}\kern-.125emX}}
\begin{document}
\title{Winter Wheat Crop Yield Prediction on Multiple Heterogeneous Datasets using Machine Learning\\
\thanks{Origin Enterprises and CONSUS}
}

\author{\IEEEauthorblockN{Yogesh Bansal}
\IEEEauthorblockA{\textit{School of Computer Science} \\
\textit{University College Dublin}\\
Dublin, Ireland \\
yogesh.bansal@ucdconnect.ie}
\and
\IEEEauthorblockN{David Lillis}
\IEEEauthorblockA{\textit{School of Computer Science} \\
\textit{University College Dublin}\\
Dublin, Ireland \\
david.lillis@ucd.ie}
\and
\IEEEauthorblockN{Tahar Kechadi}
\IEEEauthorblockA{\textit{School of Computer Science} \\
\textit{University College Dublin}\\
Dublin, Ireland \\
tahar.kechadi@ucd.ie}
}

\maketitle

\section{Abstract}
Winter wheat is one of the most important crops in the United Kingdom, and crop yield prediction is essential for the nation's food security. Several studies have employed machine learning (ML) techniques to predict crop yield on a county or farm-based level. The main objective of this study is to predict winter wheat crop yield using ML models on multiple heterogeneous datasets, i.e., soil and weather on a zone-based level. Experimental results demonstrated their impact when used alone and in combination. In addition, we employ numerous ML algorithms to emphasize the significance of data quality in any machine-learning strategy.

\begin{IEEEkeywords}
agro climatic indices, heterogeneous datasets, winter wheat yield prediction, soil and weather data
\end{IEEEkeywords}

\section{Introduction}
To make better economic and strategic decisions, farmers rely on accurate crop yield estimates. Crop yield is the amount of crop produced after harvesting, and it is measured in metric tons per hectare~\cite{zhang2002precision}. These decisions are helpful to agricultural stakeholders in taking prompt, well-informed decisions to solve crop-production issues as they arise. To satisfy rising food needs, crop production must be maximized, and farmers and agronomists must be able to predict yield while the crop is developing in order to make informed decisions regarding their inputs (growth regulators, fertilizers, etc.), which are costly. These elements may aid in identifying problems. If, for example, the yield is lower than expected, it could mean that factors, like compaction, are at play that are not shown by the statistics. These things might not be in the data because they are hard or expensive to get. 

Various factors~\cite{zhang2002precision} that can affect crop yield include the following: Agro-climatic indices are often used to show the link between weather and crop growth. These are obtained from direct weather data (temperature, precipitation, wind speed, etc.). Soil nutrients such as nitrogen (N), phosphorus (P), potassium (K), calcium carbonate (CaCO\textsubscript{3}), and magnesium (Mg) are critical for crop growth. Soil physical properties like soil texture, soil type, and stone content present in the soil affect crop production as well. Soil chemical qualities such as organic matter and pH may impact agricultural productivity, since low or high pH interferes with plants' capacity to absorb nutrients, thereby preventing or limiting crop development.

Predicting crop yield as accurately as possible is a major agricultural issue and a major challenge, as crop production is a non-linear process, owing to several complex factors~\cite{zhang2002precision}. Digital agriculture~\cite{klerkx2019review}, also known as smart farming~\cite{wolfert2017big, balakrishnan2016crop}) refers to tools that aid in the collection, storage and analysis of agricultural data on a digital platform~\cite{chergui2020impact}. It includes the use of various technologies like cloud computing (an on-demand computing system where any amount of data can be stored), artificial intelligence (simulation of human intelligence in machines), ML (an application of artificial intelligence that provides a system with the ability to automatically learn), big data~\cite{ngo2020crop} and sensors. Once high-quality data is gathered, these technologies help provide real-time insights for informed decisions, which can be crucial in solving crop production issues as they arise.

ML models are widely used in crop management systems such as yield prediction because of their ability to handle both linear and non-linear real-world data~\cite{liakos2018machine}. The accuracy of ML algorithms' predictions depends a lot on the quality of the data, the way the model is represented, the input features, and the yield, which is the target variable. Any ML model will be less able to make a prediction if the data has noisy values, wrong data, or outliers. The objective of this study is to assess the data quality using an incremental technique. We start with soil data and then add weather data to see if there is any improvement in crop yield predictions using a set of ML algorithms.

The paper is structured as follows: Section~\ref{sec:literature} presents related literature followed by data description in Section~\ref{sec:data_description}. Section~\ref{sec:data_pre-processing} gives details about preprocessing, followed by "experiment setup" and "results" in Section~\ref{sec:exp_setup} and~\ref{sec:results} respectively. Section~\ref{sec:discussion} finally presents some discussion and concludes the study.

\section{Literature Review} \label{sec:literature}
This covers the current literature on data-driven crop yield prediction using multiple types of datasets, such as soil and weather.
In \cite{van2020crop}, the authors provided a useful overview of corp datasets, models, and evaluations that are commonly used. It states that most researchers work with soil data (which includes soil nutrients and yield data) and weather data (such as temperature, rainfall, humidity, and solar radiation), as is done in this study. Most of the time, linear, ensemble, and boosting models are used to predict crop yields with ML. RMSE, MAE, and R-squared are the most commonly used evaluation metrics for evaluating the yield forecasting performance of the ML models. MAE is the mean absolute error between the real data and the predicted data. In discussions with agronomists, MAE is the most intuitive evaluation metric for them to interpret because it is expressed in metric tons per hectare. Therefore, MAE has been used in this study to evaluate the yield forecasting performance of the ML models presented in Section~\ref{sec:results}.

In another study~\cite{jeong2016random}, a random forest (RF) model was tested for wheat yield prediction, with multiple linear regression (MLR) used as the benchmark for comparison. RF was found to outperform MLR in all performance statistics. Another study~\cite{gonzalez2014predictive} compares MLR, Decision Tree (DT), Multi-Layer Neural Networks, Support Vector Regression (SVM), and K-Nearest Neighbor (KNN) for crop yield prediction on 10 crop datasets and finds that DT gives the lowest errors and is an appropriate tool for large-scale crop yield prediction.

In one study~\cite{nigam2019crop}, scientists examined the performance of RF, XGBoost, and KNN for crop yield prediction with rainfall and temperature data and found that RF was best for yield predictions. Another study conducted by~\cite{wang2020winter} employed RF, Support Vector Machine (SVM), Least Absolute Shrinkage and Selection Operator (Lasso), and neural models on soil and weather data for winter wheat yield prediction and concluded that neural models outperform other models. More relevant studies were done by~\cite{han2020prediction,cai2019integrating}, demonstrating RF as a better ML predictor than other ML models. Other studies~\cite{balakrishnan2016crop, huang2006hybrid} applied ensemble ML models, compared them with an individual ML model and concluded that ensemble ML models performed better than a single ML model. In relation to data pre-processing, Ngo et. al~\cite{ngo2019predicting} used the nearest fields to fill in the missing values in soil nutrients like pH.

More research findings demonstrate the significance of combining several data sources, such as weather and satellite data, for crop yield prediction as they are much more effective in capturing yield variability than when models are based on individual data source~\cite{cao2020identifying,wang2020winter,wang2020combining}. In other works~\cite{mathieu2018assessment,rasmussen2018climate}, authors utilize and compare agro-climatic indices to determine the weather variability for corn yield. Based on adverse weather indices, researchers evaluate a variety of bad weather circumstances that might affect crop productivity in a separate study~\cite{harkness2020adverse}. The motivation for the present study emerged from the yield-limiting factors that impact crop yield.
\FloatBarrier

\section{Data Description} \label{sec:data_description}
The data sources used in this study consist of soil and weather datasets covering a large number of farms for a number of years. A farm is divided into several fields, and a field is further divided into zones. A zone is defined as a sub-region within a field that has similar characteristics in terms of crop management~\cite{van2020crop}. Let $Z$ be a set of zones: $Z = \{z_1, z_2, \ldots, z_n\}$. For each zone $z_i$ by year, the following features are grouped into two categories: soil data and weather data. The soil information has been fetched from CONTOUR\footnote{CONTOUR is a soil dataset that gives information about the soil physical and chemical properties of a farm, field, and zone along with the zone boundaries, latitude, and longitude, provided by Origin Enterprises: the industry partner in the CONSUS project \url{https://contour.ag-space.com/}} and the weather data has come from the Iteris ClearAg weather dataset\footnote{Iteris ClearAg provides weather data for farms with a spatial resolution of 1km}. 

\begin{figure}
    \centering
    \includegraphics[width=0.5\textwidth]{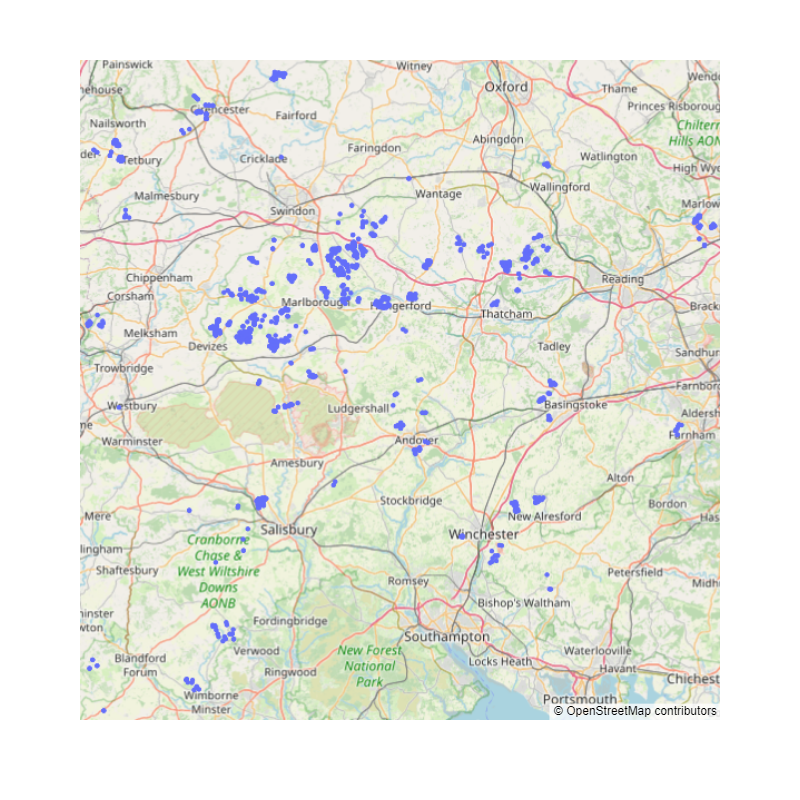}
    \caption{Study Locations in the United Kingdom}
    \label{fig:latlon}
\end{figure}

Figure~\ref{fig:latlon} shows various locations in the United Kingdom for which this study has been conducted. Some of the locations are distant from those that are nearest to one another.

The soil data contains information about the soil tests conducted in the farming zones. These soil tests are infrequent (typically every 3–4 years) because they are expensive and also because the values do not vary substantially over relatively short periods of time. Since we are using a time frame of one year per instance, we assume that the soil properties do not meaningfully change in that time and so treat them as constants. If a zone doesn't have a soil test in a given year, the most recent test done in the same zone in the year before is used to map the soil. However, weather data is a collection of a series of time points. An instance $I$ in the dataset can be represented as follows \begin{equation}
    I = [Year, ZoneID, S_x, W_y, Yield]
\end{equation}
where $S_x$ represents x soil data features and $W_y$ represents y weather data features. In other words, it can be represented as 
\begin{equation}
    I = [Year, ZoneID, S_1, \ldots, S_x, W_1, \ldots, W_y, Yield]
\end{equation}
where $I$ is represented by $S_x$ (where $x$ is the total number of soil data variables), $ZoneID$, and $W_y$ (where $y$ is the total number of weather variables used in the study).

\begin{figure}
    \centering
    \includegraphics[width=0.46\textwidth]{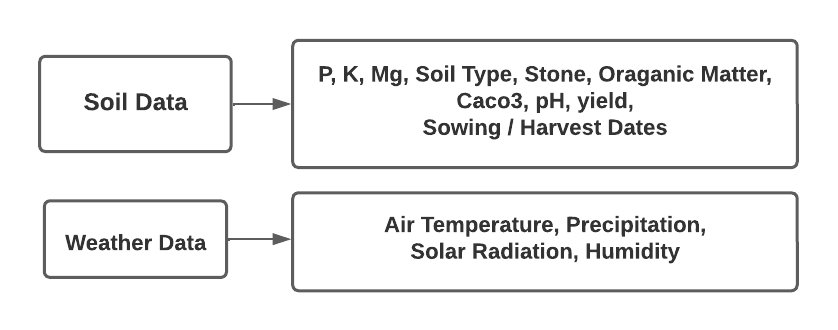}
    \caption{Data Description of a Zone}
    \label{fig:DPP}
\end{figure}

Figure~\ref{fig:DPP} shows soil and weather data attributes used in this study from 2013 to 2018. It comprises soil nutrients (P, K, Mg), physical properties (soil type, stone content), chemical properties (organic matter, CaCO\textsubscript{3}, pH), and the yield for a zone along with sowing and harvest dates. The numerous soil type classifications in the dataset are shallow, medium, deep clay, and deep fertile; stone content is stoneless, low, moderate, high, and gravel; organic matter is low, moderate, and very high; and CaCO\textsubscript{3} is slightly calc, calc, extremely calc, and potentially acidic. In addition, the weather data comprises air temperature, precipitation, solar radiation, and humidity. The total number of zones, mean yield and standard deviation of yield in each year are shown in table~\ref{tab:zcount}.
\begin{table}[hbt]
\caption{Zone count and average yield for each year} \label{tab:zcount}
\begin{center}
\begin{tabular} { | c | c | c | c | c |} 
\hline
\textbf{Years} & \textbf{Total Zones} & \textbf{Mean Yield (t/h)} & \textbf{Std Dev} \\ \hline
    2013 & 359 & 8.99 & 1.86 \\ \hline
    2014 & 335 & 10.78 & 1.61 \\ \hline
    2015 & 362 & 11.71 & 1.36 \\ \hline
    2016 & 221 & 9.94 & 1.42 \\ \hline
    2017 & 331 & 10.24 & 1.79 \\ \hline
    2018 & 264 & 9.36 & 1.75 \\ \hline
\end{tabular}
\end{center}
\end{table}

\begin{figure}[h]
    \centering
    \includegraphics[width=0.46\textwidth]{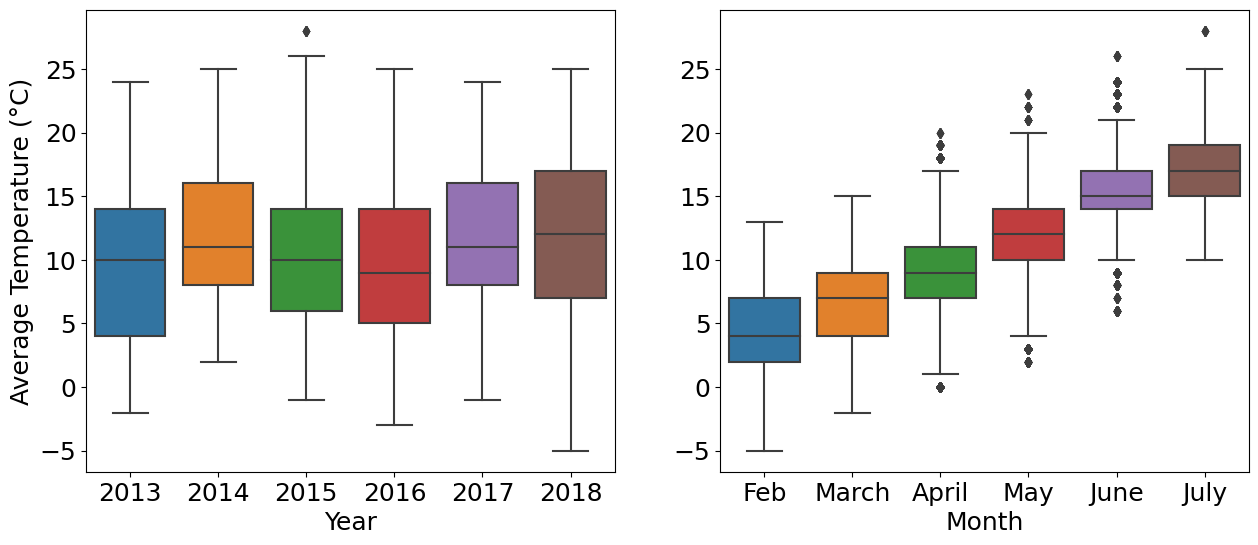}
    \caption{Trend of Average Temperature by Year and Seasonality by Month}
    \label{fig:Temperature}
\end{figure}

\begin{figure}[h]
    \centering
    \includegraphics[width=0.5\textwidth]{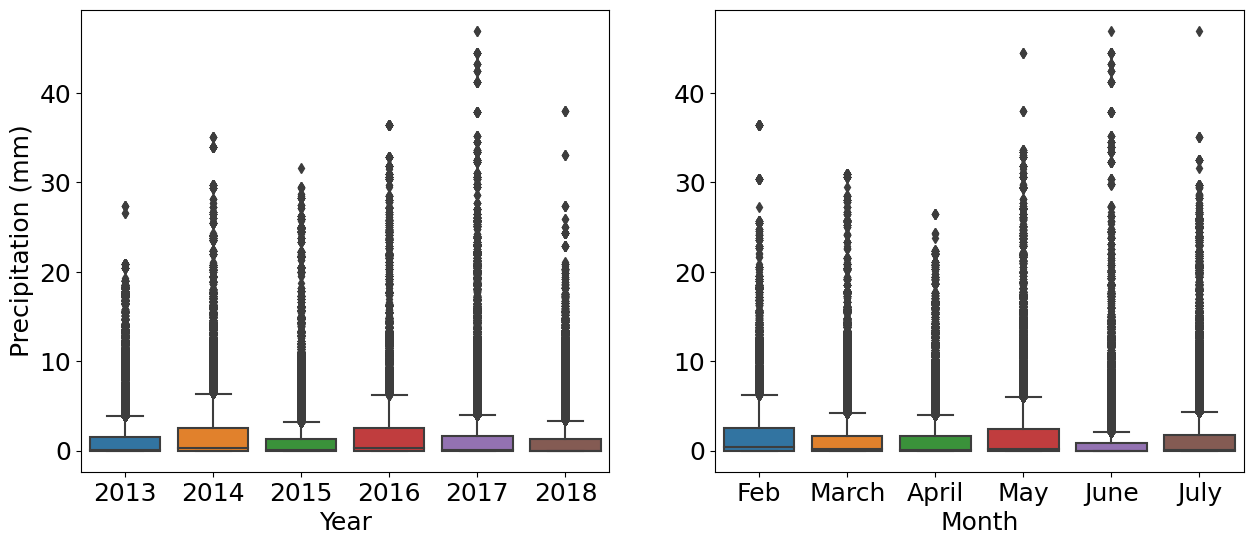}
    \caption{Trend of Precipitation by Year and Seasonality by Month}
    \label{fig:Precipitation}
\end{figure}

\begin{figure}[h]
    \centering
    \includegraphics[width=0.5\textwidth]{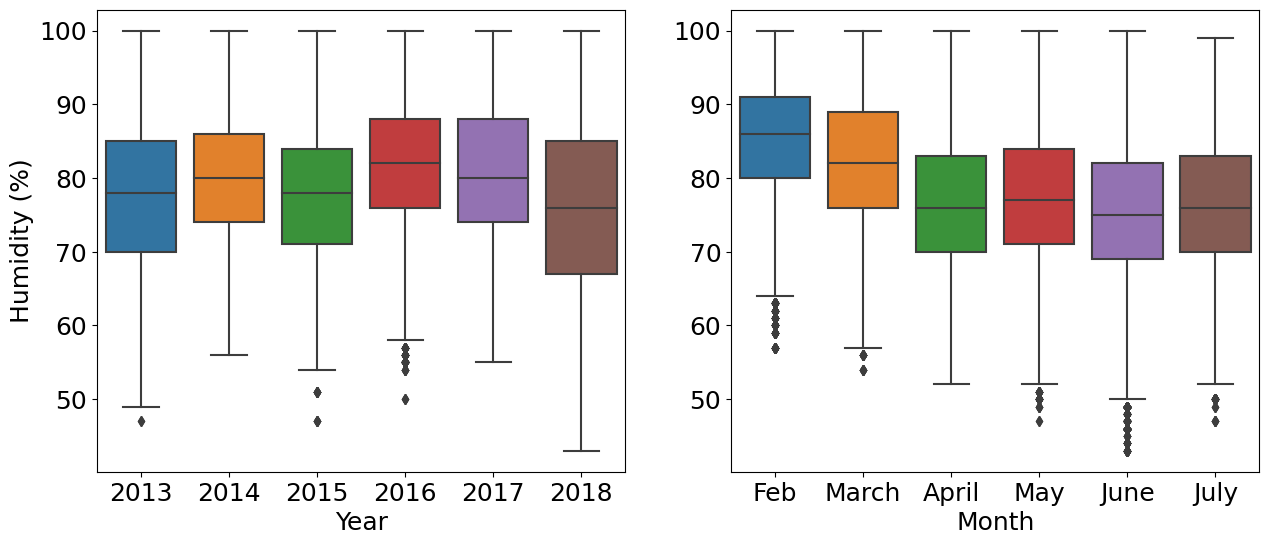}
    \caption{Trend of Humidity by Year and Seasonality by Month}
    \label{fig:Humidity}
\end{figure}

\begin{figure}[h]
    \centering
    \includegraphics[width=0.5\textwidth]{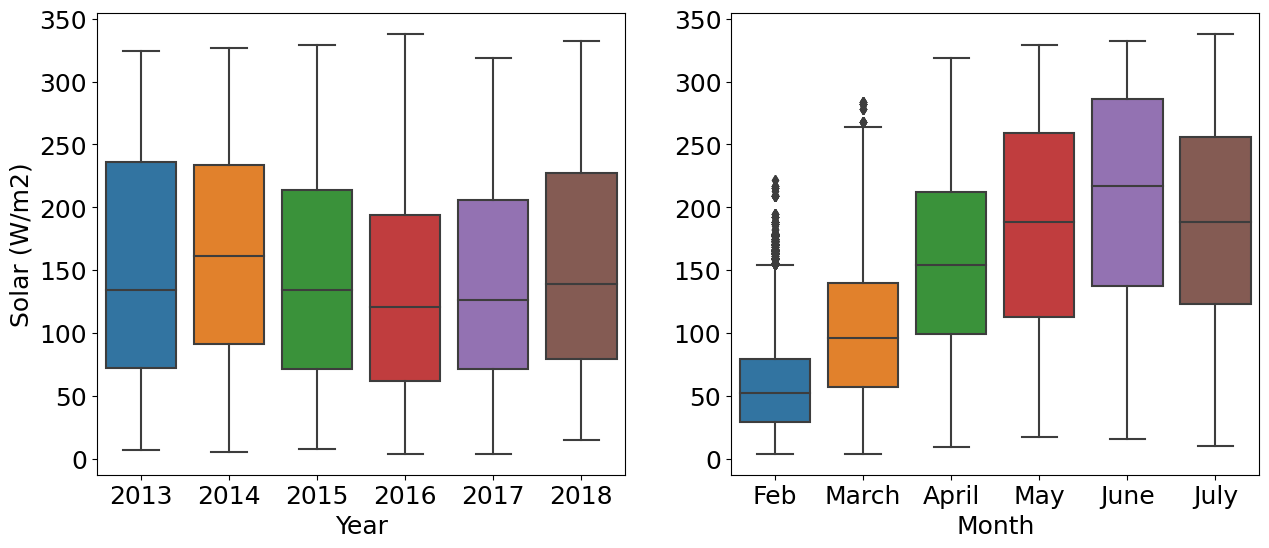}
    \caption{Trend of Solar Radiation by Year and Seasonality by Month}
    \label{fig:Solar}
\end{figure}

Box plots are used to show the yearly trend and the monthly seasonality of weather data. This helps to understand the distribution, outliers, mean, median, and variance of weather data. As only the growth period months that are significant for winter wheat are examined in this study, the corresponding months are shown in the Figures~\ref{fig:Temperature}, \ref{fig:Precipitation}, \ref{fig:Humidity}, and~\ref{fig:Solar}. 

\FloatBarrier

\FloatBarrier

The fact that soil data for a zone was only available every three to four years while weather data was available every day shows how different the original datasets were. These have been examined, preprocessed, and brought into a common format, ready for further analysis by ML models. So, this study was done on multiple datasets to predict the crop yield of winter wheat at the zone level, where each zone has its own soil and weather data.
\FloatBarrier

\section{Data Preprocessing} \label{sec:data_pre-processing}
The original soil and weather data contain many inconsistencies, outliers, noise, and missing values. It is necessary to preprocess it before analysis. The data preprocessing procedure begins with identifying and correcting data errors, filling in missing values, applying feature engineering techniques to extract new features, integrating data, and finally transforming and filtering the data to prepare it for modelling purposes.
\FloatBarrier

\begin{figure}[h]
    \centering
    \includegraphics[width=0.45\textwidth]{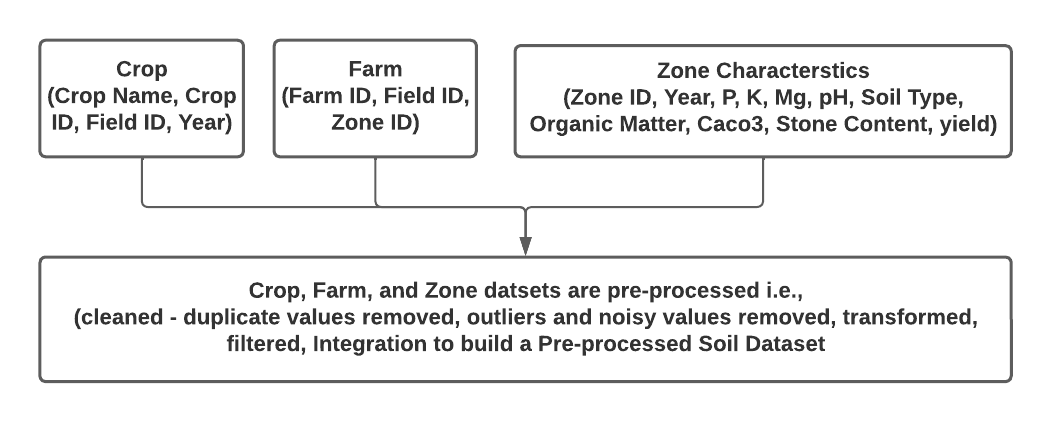}
    \caption{Soil Data Preprocessing and Integration}
    \label{fig:spp}
\end{figure}

Figure~\ref{fig:spp} depicts several datasets that have been preprocessed and then merged to produce a master soil dataset. Winter wheat was filtered from the original soil dataset that also included many other crops. A few duplicate soil testing and yield values for the same zone and year were eliminated. With the assistance of domain specialists in agriculture, several soil nutrient and yield data values were determined to be incorrect and/or impossible. Those are most likely the result of human error. These were removed from the soil data collection. Soil ordinal data, such as soil type, stone content, organic matter, and CaCO\textsubscript{3} were converted into numerical characteristics using ordinal encoding. After the crop, farm, and zone datasets have been preprocessed, they are combined to generate a soil dataset.

\begin{figure}[h]
    \centering
    \includegraphics[width=0.45\textwidth]{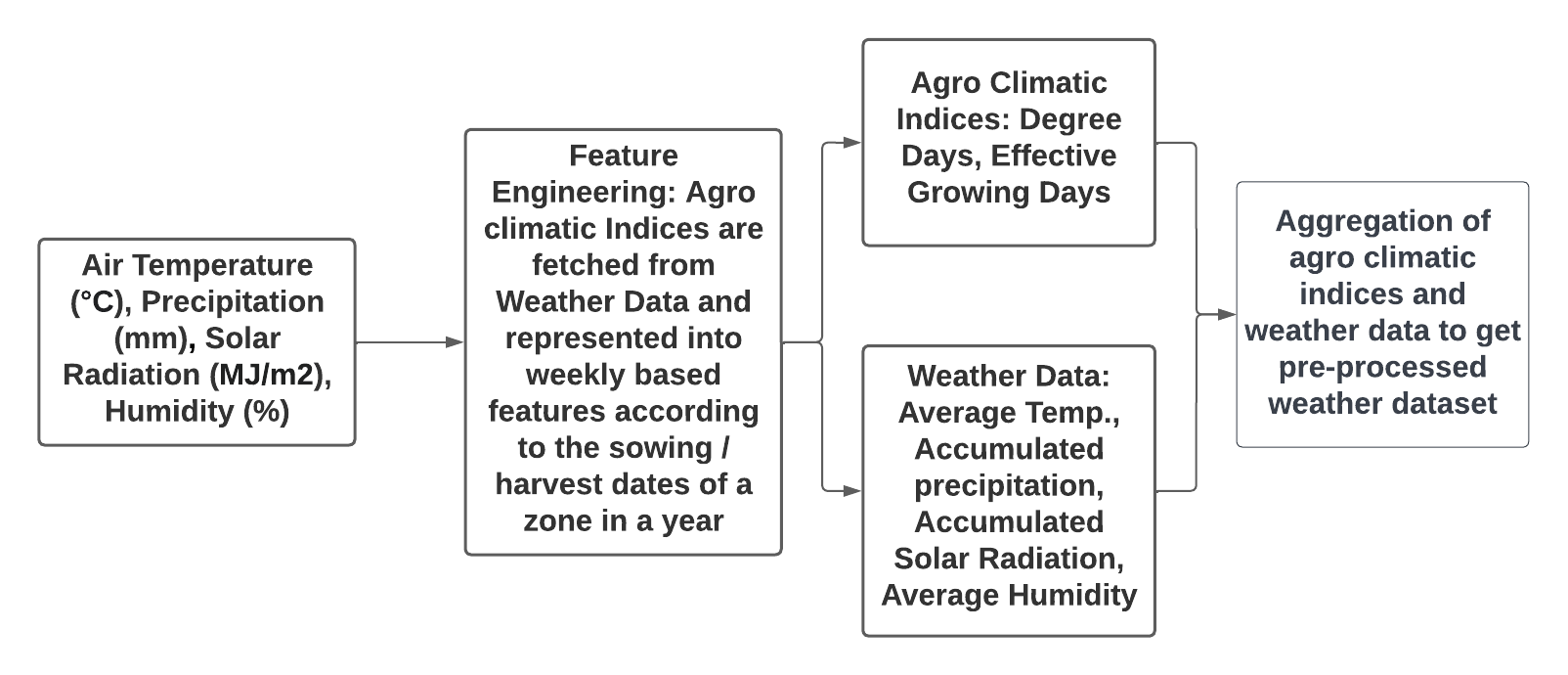}
    \caption{Weather Data Preprocessing}
    \label{fig:dpp}
\end{figure}

\FloatBarrier
Figure~\ref{fig:dpp} illustrates the pre-processing stages used for weather data. Temperature, precipitation, solar radiation, and humidity are employed in this study as weather variables. The goal of feature engineering is to find new, useful characteristics and use them in modelling. Since we are interested in representing time series data in a weekly-based analysis, pre-processed weather data is used for each week of all years investigated (2013–2018).
\FloatBarrier
The average temperature ($T_{avg}$) is calculated by averaging the temperature (T) over a week. As one of the most crucial weather factors affecting crop yield, according to~\cite{mathieu2018assessment}, we used temperature in this study to derive two agro-climatic indices: degree days and effective growing days.

Degree days ($DD_{sum}$) are calculated as the maximum of 0 and the average of the maximum daily temperature ($T_{max}$) and the minimum daily temperature ($T_{min}$), summed over a week. The total number of days in a week when the average temperature is greater than 5°C is referred to as effective growing days ($EGD_{total}$). The accumulated precipitation ($AP_{sum}$) is calculated by adding precipitation (P) over a week. Solar radiation (SR) is summed over a week to get accumulated solar radiation ($SR_{sum}$). Humidity (H) is averaged over the week to get average humidity ($H_{avg}$). The formulas for weather data are listed below.

\begin{figure}[h]
    \centering
    \includegraphics[width=0.45\textwidth]{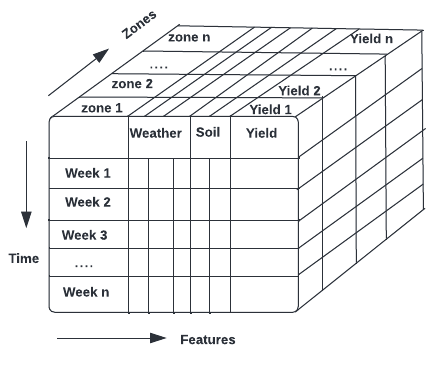}
    \caption{3D Cube showing features after data integration}
    \label{fig:3d_cube}
\end{figure}

\begin{itemize}
    \item $T_{avg} = mean((T), week)$
    \item $DD_{sum} = sum(max(0, (T_{max} + T_{min})/2), week)$
    \item $EGD_{total} = count((T_{avg} > 5^\circ C), week)$
    \item $AP_{sum} = sum((P), week)$
    \item $SR_{sum} = sum((SR), week)$
    \item $H_{avg} = mean((H), week)$
\end{itemize}    
After preprocessing soil and weather data, the data integration step begins. Figure~\ref{fig:3d_cube} illustrates the week-based representation of weather and soil data features for each zone after integrating preprocessed soil and weather data. The x-axis represents features, i.e., weather and soil. The y-axis represents time i.e., 1\textsuperscript{st} week, 2\textsuperscript{nd} week, ..., n\textsuperscript{th} week. The z-axis represents zones i.e., 1\textsuperscript{st} zone, 2\textsuperscript{nd} zone, ..., n\textsuperscript{th} zone.
\FloatBarrier

\begin{figure}[h]
    \centering
    \includegraphics[width=0.47\textwidth]{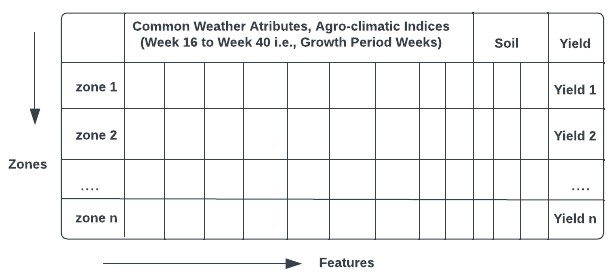}
    \caption{2D Matrix showing features after data transformation, filtering}
    \label{fig:2d_cube}
\end{figure}

The last step is to transform and filter the data, which makes the combined weather and soil data ready to use for modelling purposes. Figure~\ref{fig:2d_cube} depicts each instance based on common weather attributes and agro-climatic indices for growth period weeks and soil data, as well as yield for each zone. 
\FloatBarrier
For modelling purposes, not all weeks are essential for winter wheat production. Only growth period weeks that are important for winter wheat yield production are included in the study, i.e., weeks beginning with sowing dates through week 16 are omitted, and weeks from 17 through 40 are included (where week 1 is the week where the crop was sown). In practice, this typically covers a period between February and July.
\FloatBarrier

\section{Experiments Description Setup} \label{sec:exp_setup}
\begin{itemize}
    \item  This study is predicated on the understanding that weather has a significant impact on crop yield. Consequently, the primary aim is to demonstrate this through data analytics (performing the analytics on real-world datasets). In this work, we apply a series of ML algorithms to predict winter wheat crop yield using a) just soil data and b) integrated soil and weather data. We design a set of experiments to observe whether there is a significant effect and assess the data preprocessing efficiency developed for these datasets.
    
    The weather data is presented in two forms: common weather attributes (i.e., temperature, humidity, precipitation, solar radiation), and agro-climatic indices (i.e., degree days, effective growing days).
    \item We carried out the ML model evaluation not only by observing the error change but also by its significance. For instance, we evaluate if there is a significant difference between the absolute errors from soil to integrated soil and weather data.
\end{itemize}
The data utilized in this study covers the years 2013 to 2018. We train a collection of ML algorithms using data from 2013 to 2017 and then test them on the 2018 dataset. Utilized are several regression models (Decision Tree, Support Vector, Random Forest, Extra Trees, LightGBM, Gradient Boosting). MAE is selected as the evaluation metric.

\section{Experiment Results} \label{sec:results}
Several experiments have been designed based on the data attributes. We started with soil attributes and then added the weather attributes. These are described in the following:

\subsection{Winter Wheat Yield Predictions on Soil Data}
\begin{figure}[h]
    \includegraphics[width=0.5\textwidth]{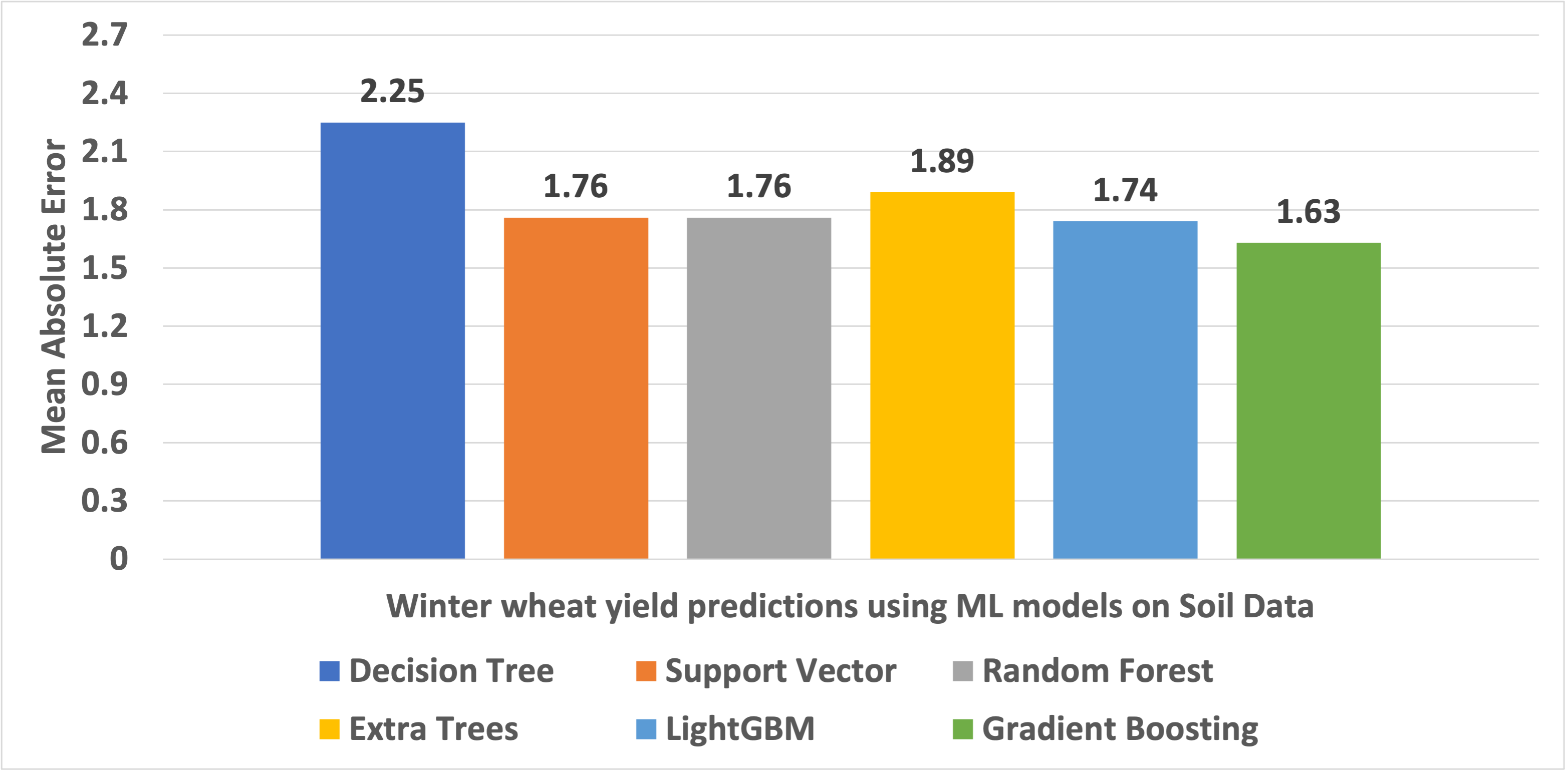}
    \caption{Yield Predictions on Soil Data}
    \label{fig:soil_2018}
\end{figure}

Table~\ref{tab:mae-soil} and Figure~\ref{fig:soil_2018} depict MAE, z-score, and p-values on the evaluation year 2018 after training the soil data for growth period weeks from 2013-2017. To compare the ML models within soil data, we carried out statistical significance tests.
\begin{table}[hbt]
\caption{Yield Predictions on Soil Data}
\label{tab:mae-soil}
\begin{center}
\centering
\begin{tabular} { | c | c | c | c |}
\hline
    Models & Soil Data & z score & p value \\ \hline
    Decision Tree & 2.25 & 2.10 & 0.017 \\ \hline
    Support Vector & 1.76 & -0.32 & 0.624 \\ \hline
    Random Forest & 1.76 & -0.32 & 0.624 \\ \hline
    Extra Trees & 1.89 & 0.32 & 0.372 \\ \hline
    LightGBM & 1.74 & -0.41 & 0.661 \\ \hline
    Gradient Boosting & \textbf{1.63} & -0.96 & 0.831 \\ \hline
\end{tabular}
\end{center}
\end{table}
\FloatBarrier
For each ML model, the z-score is calculated using the mean and standard deviation of all ML models, and the p-value is calculated using a standard normal distribution. It is noted that although the ML models slightly differ in their predictions, with gradient boosting having the lowest MAE of 1.63 t/h, these differences are not significant.
\FloatBarrier

\subsection{Winter Wheat Yield Predictions on Integrated Soil and Weather Data}
\begin{figure}[h]
    \includegraphics[width=0.5\textwidth]{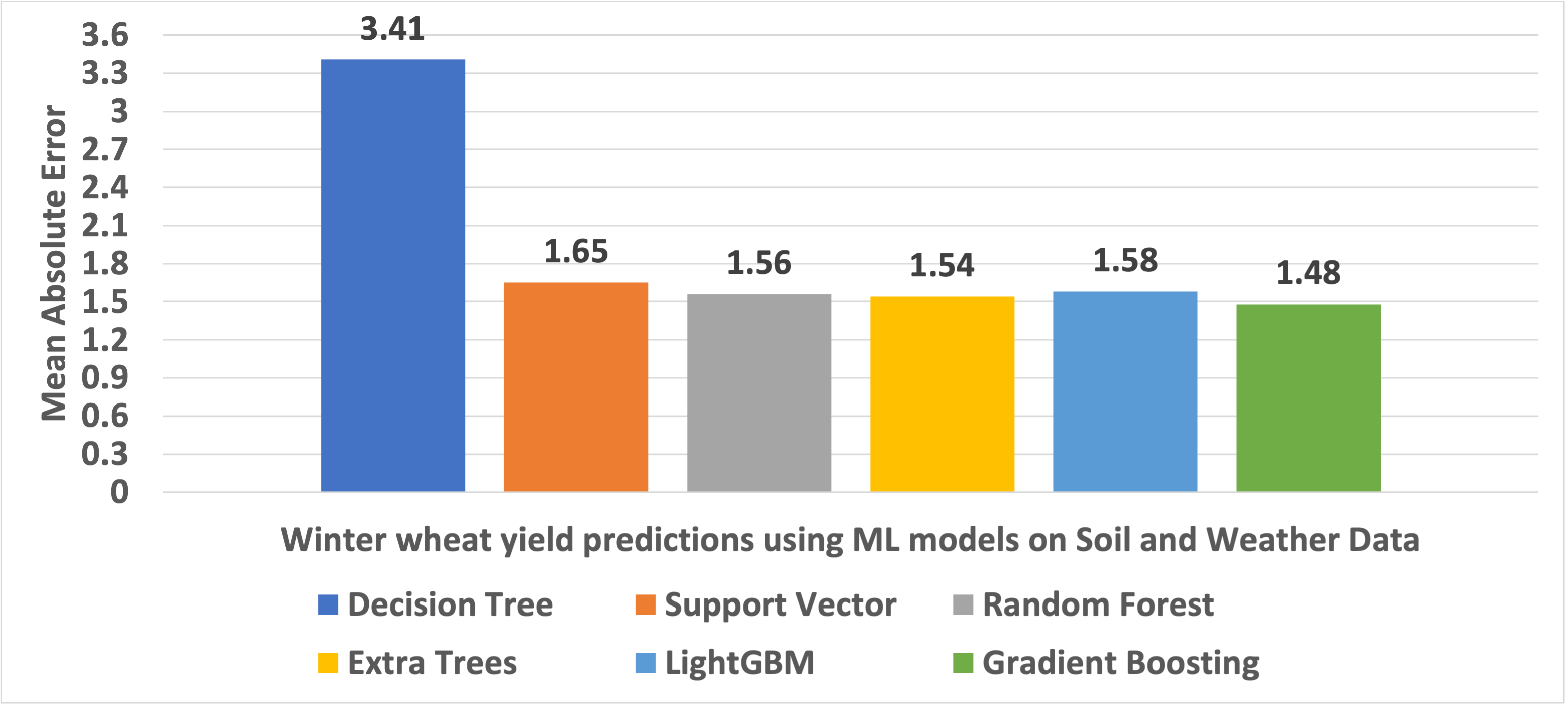}
    \caption{Yield Predictions on Integrated Soil and Weather Data}
    \label{fig:soil_weather_2018}
\end{figure}

Table~\ref{tab:mae-soil_weather} and Figure~\ref{fig:soil_weather_2018} show MAE, z-score, and p-values on the same evaluation year 2018 after training the integrated soil and weather data for growth period weeks from 2013-2017. Here likewise, statistical significance tests are done to compare the ML models' performance. Results suggest a similar trend, i.e., the ML models vary somewhat in their predictions; gradient-boosting has the lowest MAE of 1.48 t/h, but these variations are not statistically significant.

\begin{table}[hbt]
\caption{Yield Predictions on Integrated Soil and Weather Data}
\label{tab:mae-soil_weather}
\begin{center}
\centering
\begin{tabular} { | c | c | c | c|} 
\hline
    Models & Soil and Weather Data & z score & p value \\ \hline
    Decision Tree & 3.41 & 2.26 & 0.012 \\ \hline
    Support Vector & 1.65 & -0.28 & 0.610 \\ \hline
    Random Forest & 1.56 & -0.41 & 0.658 \\ \hline
    Extra Trees & 1.54 & -0.44 & 0.669 \\ \hline
    LightGBM & 1.58 & -0.38 & 0.648 \\ \hline
    Gradient Boosting & \textbf{1.48} & -0.52 & 0.700 \\ \hline
\end{tabular}
\end{center}
\end{table}
\FloatBarrier

\subsection{Comparison of yield predictions on Soil Data and Integrated Soil and Weather Data}
\begin{figure}[h]
    \includegraphics[width=0.5\textwidth]{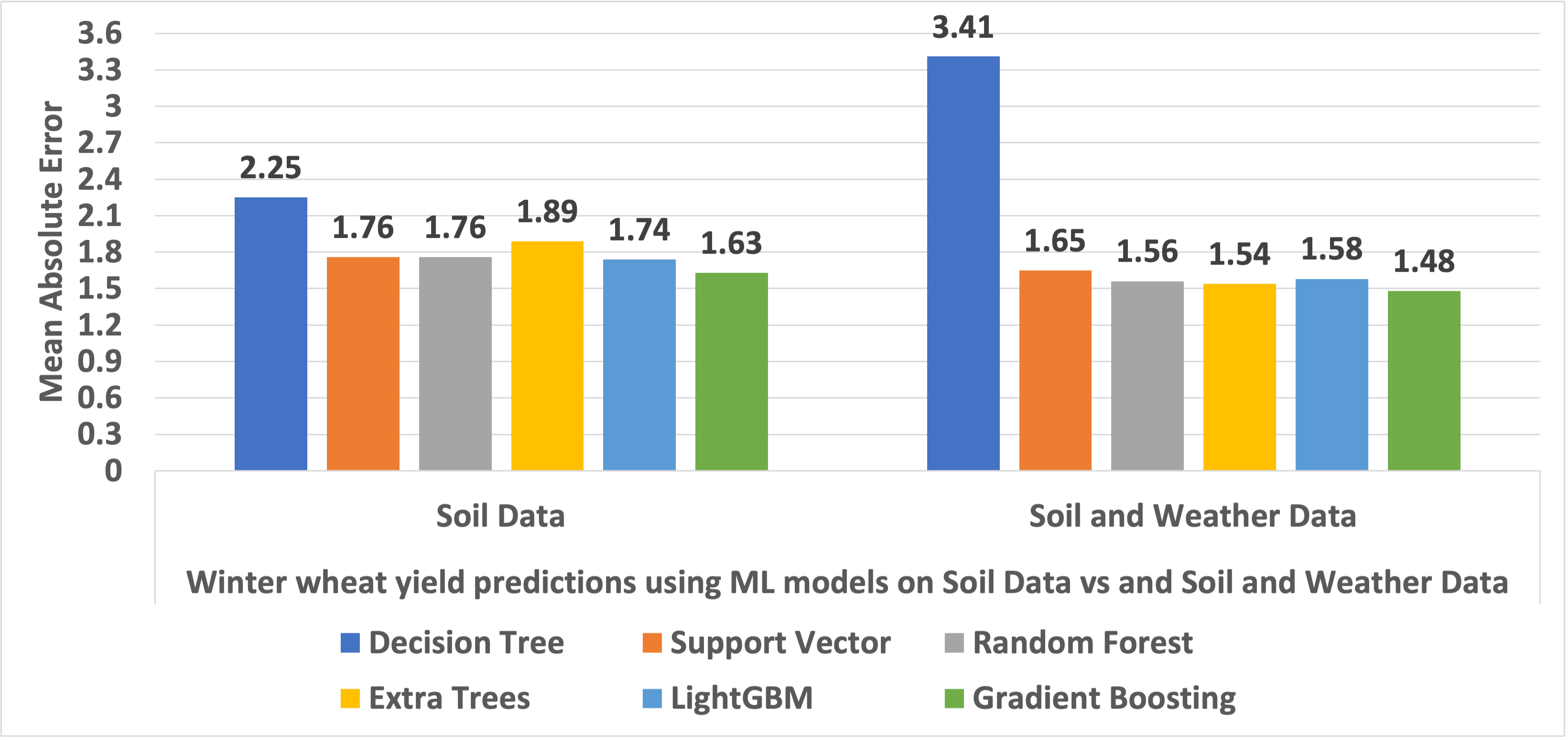}
    \caption{Comparison of Yield Predictions on Soil Data and Integrated Soil and Weather Data}
    \label{fig:soil_vs_soil_and_weather_2018}
\end{figure}

Table~\ref{tab:mae-soil_vs_soil_and_weather} and Figure~\ref{fig:soil_vs_soil_and_weather_2018} illustrate MAE, and p-values for soil and integrated soil and weather data on the same evaluation (2018) and training period (2013-2017). Since we are trying to measure whether the addition of weather data to soil data improves yield prediction, our alternative hypothesis is that the MAE of integrated soil and weather data is less than the MAE when only soil data is used. Here, p-values are calculated using a one-tailed paired t-test by comparing the absolute errors on soil and integrated soil and weather data for each ML model. 

It is noted from the results that p-values for implemented ML models are below the $5\%$ threshold, and thus the alternative hypothesis can be accepted. Consequently, it can be said that there is a considerable difference between predictions based on soil data alone and predictions based on soil plus weather data, and the performances of ML models may be compared. 

Results state that linear models like decision trees do not perform well when weather data is added; however, non-linear models (e.g., ensemble models and boosting algorithms) improve the yield predictions. Gradient Boosting captured yield variability most effectively among the ML models and was able to demonstrate an improvement from 1.63 t/h to 1.48 t/h when the weather data was added to the original soil-only dataset.
\begin{table}[hbt]
\caption{Comparison of Yield Predictions on Soil Data and Integrated Soil and Weather Data}
\label{tab:mae-soil_vs_soil_and_weather}
\begin{center}
\centering
\begin{tabular} { | c | c | c | c |}
\hline
    Models & Soil Data & Soil and Weather Data & p value \\ \hline
    Decision Tree & \textbf{2.25} & 3.41 & 6e$^{-13}$ \\ \hline
    Support Vector & 1.76 & \textbf{1.65} & 0.011 \\ \hline
    Random Forest & 1.76 & \textbf{1.56} & 0.002 \\ \hline
    Extra Trees & 1.89 & \textbf{1.54} & 1e$^{-5}$ \\ \hline
    LightGBM & 1.74 & \textbf{1.58} & 0.001 \\ \hline
    Gradient Boosting & 1.63 & \textbf{1.48} & 0.027 \\ \hline
\end{tabular}
\end{center}
\end{table}
\FloatBarrier

\section{Discussion and Conclusion} \label{sec:discussion}
This study intends to investigate whether the addition of weather data to soil data is helpful in making winter wheat crop yield predictions, based on the data collected from a wide range of fields in the UK. The data is heterogeneous and contains significant missing values, noise, and outliers. After preprocessing the data, the goal is to assess its validity in predicting crop yield. 

We use several learning algorithms on this dataset. We first observe that these algorithms return the same error level. Second, we observe that while the error was improved between soil data and soil and weather data combined, it is not at the level that it should be, meaning that the data is not representative enough to highly reflect the effect of weather conditions on crop yield for many reasons, including a) The data is still very noisy (for instance, the soil data within one year is supposed to be the same, which is a gross approximation); b) The dataset is small and not representative enough, and c) Other dimensions were not considered in this study due to a lack of data and may have a very significant impact on yield. 

One can notice that linear learning models are not performing well on this application, which was well predicted as the correlations between the soil, weather data, and yield are not linear. In general, non-linear ML models demonstrated enhanced crop yield prediction ability; gradient boosting with the lowest MAE on soil data and on the integration of soil and weather data.

This mixed-type dataset has geographical and temporal dimensions, with temporal dimensions outweighing spatial dimensions. Consequently, it is crucial to examine them using time-series learning methods, the focus of our subsequent work.

\FloatBarrier

\section{Acknowledgement}
This research is funded under the SFI Strategic Partnerships Programme  (16/SPP/3296)
and is co-funded by Origin Enterprises Plc.

\bibliographystyle{IEEEtran}
\bibliography{bibliography.bib}

\end{document}